\documentclass{svproc}
\usepackage{url}
\usepackage{cite}
\usepackage{amsmath,amssymb,amsfonts}
\usepackage{algorithmic}
\usepackage{graphicx}
\usepackage{textcomp}
\usepackage{xcolor}
\usepackage{subcaption}
\usepackage{caption}
\usepackage{float}
\usepackage[showdeletions]{color-edits}

\begin{document}
\mainmatter              
\title{Spatial-Temporal Reinforcement Learning for Network Routing with Non-Markovian Traffic
\thanks{Kin K. Leung was supported by the EPSRC grant EP/Y037243/1 and the Dstl SDS Continuation Project.}
}
\titlerunning{Spatial-Temporal RL}  
%

\author{Molly Wang\inst{1} \and Kin K. Leung\inst{2}}
\authorrunning{First Author et al.} 
%
\tocauthor{First Author, Second Author, Third Author and Fourth Author}
\institute{Department of Computing, Imperial College London, UK,\\
\email{jw923@ic.ac.uk}
\and
Departments of Electrical \& Electronic Engineering, and Computing, Imperial College London,\\
Exhibition Rd, South Kensington, London SW7 2AZ}

\maketitle              

\begin{abstract}
Reinforcement Learning (RL) has been widely used for packet routing in communication networks, but traditional RL methods rely on the Markov assumption that the current state contains all necessary information for decision-making. In reality, internet traffic is non-Markovian, and past states do influence routing performance. Moreover, common deep RL approaches use function approximators, such as neural networks, that do not model the spatial structure in network topologies. To address these shortcomings, we design a network environment with non-Markovian traffic and introduce a spatial-temporal RL (STRL) framework for packet routing. Our approach outperforms traditional baselines by more than 19\% during training and 7\% for inference despite a change in network topology.
\end{abstract}
\section{Introduction}
Rapid growth in network size and traffic volume, driven by social media, video streaming, and 5G networks, has created new challenges for routing optimization in communication networks. One significant issue is inter-domain traffic engineering or Wide-Area Network (WAN) optimization, which involves optimizing traffic routing across multiple domains. These problems are often NP-hard due to the combinatorial nature of routing paths, making analytical methods difficult to apply \cite{zheng2015minimizing}. Traditional machine learning approaches to traffic routing often require labeled data (e.g., next hop, path), which may not always be available for large-scale networks due to operational and privacy concerns \cite{ml_routing}.

Reinforcement learning (RL) offers a model-free approach, eliminating the need for a mathematical model or details of the network environment. Instead, RL learns optimal routing policies through interactions with the environment in the form of rewards. Deep reinforcement learning (DRL) improves RL by using deep neural networks to approximate routing policies, particularly suitable for complex communication network environments where traditional tabular RL methods become computationally infeasible. Recent studies have successfully applied DRL techniques to address challenging network routing problems \cite{drl_sdn1, drl_sdn2}.

Despite their popularity, DRL methods rely on the Markovian assumption, which is often violated by the non-Markovian nature of real-world Internet traffic \cite{non_markovian}. Therefore, we use a type of recurrent neural network to capture such temporal patterns. Moreover, DRLs often use fully connected feedforward neural networks to make decisions, which do not model the spatial relationships (e.g., neighboring connections) in the network topology. To address this, we use a form of Graph Neural Network to reflect the spatial relationships.

Therefore, this paper proposes a Spatial-Temporal Reinforcement Learning (STRL) framework to capture both spatial and temporal relationships in communication networks using Graph Attention Networks (GATs) and Gated Recurrent Units (GRUs). In particular, we apply the STRL approach to solving the network routing problem. The contributions of this paper are:

\begin{itemize}
\item Propose an improved STRL approach to capture non-stationary patterns in network traffic utilizing GRUs with an attention mechanism.
\item Capture spatial relationships using GATs.
\item Demonstrate the importance of jointly considering spatial-temporal dynamics for better routing decisions
\item Demonstrate the robustness of the proposed method against change of network topology, which is not included in the training data.
\end{itemize}
\section{Related Works}
Periodic Markov Decision Processes (pMDPs) have been proposed to capture periodic patterns in internet traffic \cite{chen2023periodic}, but they remain inadequate for modeling more complex non-stationary patterns. On the other hand, recurrent neural networks, such as GRUs, can capture such complex temporal patterns. Convolutional neural networks (CNNs) have been used to represent spatial structures, but are designed for grid-like Euclidean data (e.g., images), making them not suitable for networks with arbitrary topologies \cite{chen2020dynamic, anderson2021limitation}. More recently, Graph Neural Networks (GNNs), which are designed to model arbitrary network structures, have been applied in combination with time series of link traffic loads to capture temporal dynamics \cite{poularakis2020gat}. However, these approaches address link-level congestion prediction rather than the challenge of routing optimization as considered in this paper.

\section{Preliminaries}

\subsection{Reinforcement Learning}
Reinforcement learning (RL) is a machine learning approach in which an agent learns to make decisions by interacting with an environment. The key components of RL include the agent and the environment with  states (\(s\)), actions (\(a\)), rewards (\(r\)), and policy (\(\pi\)). Formally, the environment is modeled as a Markov Decision Process (MDP), defined by the tuple \(\langle \mathcal{S}, \mathcal{A}, r, \mathcal{P} \rangle\), where \(\mathcal{S}\) is the state space, \(\mathcal{A}\) is the action space, \(r\) is the reward function, and \(\mathcal{P}\) denotes the state transition probability. At each time step, the agent observes the current state \(s \in \mathcal{S}\), selects an action \(a = \pi(s)\) according to its policy \(\pi\), and receives a corresponding reward \(r\). The environment then transitions to a new state \(s' \in \mathcal{S}\) with probability \(\mathcal{P}_{ss'}^{a} = P(s'|s, a)\). The agent's goal is to learn a policy that selects actions in each state to maximize the expected sum of rewards over time.

\subsection{Deep Deterministic Policy Gradient (DDPG)}
\label{sec:DDPG}

Reinforcement learning (RL) algorithms are broadly categorized into value-based methods, which learn a value function that estimates the expected cumulative reward for each state-action pair (e.g., Q-learning, DQN), and policy-based methods, which directly learn a policy mapping states to action probabilities (e.g., REINFORCE). To take advantage of both approaches, we use the Deep Deterministic Policy Gradient (DDPG) algorithm.

DDPG offers an off-policy actor-critic framework, where “actor-critic” refers to the use of two neural networks: an actor that learns the policy (i.e., selects actions) and a critic that estimates the value function to evaluate the actor’s actions. “off-policy” means that the algorithm learns the optimal policy by making use of the data generated from a different policy called the behavior policy, to improve sample efficiency. As a result, DDPG employs four neural networks: the behavior actor (\(\mu\)) and critic (\(Q\)), as well as their target networks (\(\hat{\mu}\) and \(\hat{Q}\)). The “deep” aspect refers to parameterizing both the actor and the critic with deep neural networks. The optimization process is as follows.

We first optimize the behavior critic network, parameterized by \(\theta\), which outputs the action-value function \(Q_\theta(s, a)\), i.e., the expected cumulative reward starting from state \(s\) and taking action \(a\), and then following the current policy. The critic is trained by minimizing the mean squared error loss
\begin{equation}
L(\theta) = \mathbb{E}\left[(Q_\theta(s, a) - y)^2\right],
\label{eq:loss}
\end{equation}
where \(y\) is the \textit{target value} representing the estimated return for the next state-action pair, which is computed using the Bellman equation:
\[
y = r + \gamma \, Q_{\hat{\theta}}\big(s', \hat{\mu}_{\hat{\omega}}(s')\big),
\]
with \(s'\) denoting the next state, \(\hat{\mu}_{\hat{\omega}}\) representing the target actor network (parameterized by \(\hat{\omega}\)), which generates the action for the next state, and \(Q_{\hat{\theta}}\) denoting the target critic network (parameterized by \(\hat{\theta}\)), which evaluates the expected return of that next state-action pair. Here, \(\gamma \in [0, 1]\) is the discount factor.

Next, we optimize the behavior actor network \( \mu \), which predicts actions that aim to maximize the \( Q \)-value estimated by the behavior critic network \( Q \). The actor parameters \( \omega \) are updated through gradient ascent, with the gradient calculated as
\[
\nabla_{\omega} J(\mu) \approx \mathbb{E} \left[ \nabla_a Q_\theta(s, a) \big|_{a=\mu(s)} \nabla_{\omega} \mu(s) \right],
\]
where \( J(\mu) = \mathbb{E}_{s \sim \tau} \left[ Q_\theta(s, \mu(s)) \right] \) is the expected action-value and \(\tau\) denotes the stationary distribution of states under the current policy. The term \( \nabla_a Q_\theta(s, a) \big|_{a=\mu(s)} \) captures how the \( Q \)-value changes with respect to the action, while \( \nabla_{\omega} \mu(s) \) reflects how to adjust the actor’s parameters to maximize the expected return. In practice, the expectation is approximated using a mini-batch of size \(M\).

The target actor and critic networks are updated via exponential moving averages of the behavior networks' parameters:
\[
\hat{\omega} \leftarrow \rho \omega + (1 - \rho) \hat{\omega}, \quad \hat{\theta} \leftarrow \rho \theta + (1 - \rho) \hat{\theta},
\]
where \( \rho \in (0, 1) \) controls the update rate.


\section{Gated Recurrent Units (GRUs)}
\label{sec:GRU}
Gated Recurrent Units (GRUs) are a type of recurrent neural network designed to model temporal dependencies in time series data \cite{cho2014learning}. At each time step \(t\), the hidden state \(h^t \in \mathbb{R}^d\) summarizes information from previous steps and is updated using the current input \(x^t\) and the previous hidden state \(h^{t-1}\):
\begin{equation}
h^t = f(Wx^t + Uh^{t-1} + b),
\label{eq:update_rule}
\end{equation}
where the weight matrix \(W\) maps the input to the hidden state space, \(U\) captures the influence of the previous hidden state, \(b\) is a bias term and \(f\) is a nonlinear activation function (e.g., \texttt{tanh} or \texttt{ReLU}).

Therefore, The GRU takes as input a time series matrix \( X \in \mathbb{R}^{T \times F} \), where each row \( x^t \in \mathbb{R}^F \) represents the environment state at time \( t \) with \( F \) features. Here, \( T \) denotes the window length.

The GRU is chosen over conventional RNNs because it mitigates the vanishing gradient problem \cite{bengio1994learning} using two gating mechanisms: the update gate \( z^t \) and the reset gate \( r^t \), which regulate how much past information is retained:
\begin{align}
z^{t} = \sigma \left(W^{z}x^{t} + U^{z}h^{t-1}\right),\quad
r^{t} = \sigma \left(W^{r}x^{t} + U^{r}h^{t-1}\right),
\label{eq:gating_reset_gate}
\end{align}
where \( W^z \), \( W^r \), \( U^z \) and \( U^r \) are weight matrices. The reset gate \( r^t \) then determines how much of the previous hidden state is considered when computing the candidate hidden state:
\begin{align}
\tilde{h}^t = \tanh\left(Wx^t + r^t \odot Uh^{t-1}\right),
\label{eq:hidden_layer_output}
\end{align}
where \( W \) and \( U \) are weights for the input and hidden state. Next, the update gate \( z^t \) controls how much of the previous hidden state versus the candidate hidden state contributes to the new hidden state:
\begin{align}
h^t = z^t \odot h^{t-1} + (1 - z^t) \odot \tilde{h}^t.
\label{eq:unit_output}
\end{align}
Through this mechanism, the GRU adaptively retains, updates, or discards past information at each time step, resulting in a sequence of hidden states:
\begin{equation}
H = \{ h^1, h^2, \ldots, h^T \} \in \mathbb{R}^{T \times d}.
\label{eq:hidden_state}
\end{equation}

\subsection{Temporal Attention Mechanism}
\label{sec:attention}
However, GRUs may struggle to retain information from the distant past time steps. To address this, a temporal-attention mechanism is introduced to allow the model to focus on relevant points in the sequence using attention weights. These weights are calculated through a query key mechanism where three vectors are used: \textit{ query} (\(Q\)), which encodes what the current time step is searching for; the \textit{key} (\(K\)), which represents the characteristics of each past time step; and the \textit{value} (\(V\)), which contains the information from those time steps. 

Specifically, we first obtain \(Q\), \(K\), and \(V\) by projecting the hidden states \(H\) through the learnable parameter matrices \(W_Q \in \mathbb{R}^{d \times d_q}\), \(W_K \in \mathbb{R}^{d \times d_k}\), and \(W_V \in \mathbb{R}^{d \times d_v}\):
\begin{align}
    Q = W_Q H, \quad K = W_K H, \quad V = W_V H,
    \label{eq:self_attention_components}
\end{align}
into different representational spaces for querying, matching relevance, and aggregating information. As a result, \( Q \in \mathbb{R}^{T \times d_q} \), \( K \in \mathbb{R}^{T \times d_k} \), and \( V \in \mathbb{R}^{T \times d_v} \). Then, we calculate attention weights \( \alpha_{ij} \) which measures the relevance of the \(j\)-th past time step to the current step \(i\) as follows,
\begin{align}
\alpha_{ij} = \frac{\exp(k_j^T \cdot q_i)}{\sum_{j'=1}^{i} \exp(k_{j'}^T \cdot q_i)} \quad \text{for } j \leq i,
\label{eq:attention_weights}
\end{align}
with query \(q_i\) and key \(k_j\). Thus, more relevant past steps receive higher weights.

Then, we obtain the context vector \( h^a_t \) at time \( t \) by aggregating information from all previous time steps. That is, the context vector is computed as a weighted sum of the value vectors
\[
h^a_t = \sum\nolimits_{j=1}^{t} \alpha_{tj} \cdot v_j,
\]
Collecting the context vectors for all time steps yields the following.
\begin{align}
H^a = \left[ h^a_1, h^a_2, \ldots, h^a_T \right] \in \mathbb{R}^{T \times d_v},
\label{eq:attention_matrix}
\end{align}
which represents the attended features across the sequence.

Finally, we combine each hidden state in the original sequence \( H \) with its corresponding attended context from (\ref{eq:attention_matrix}), resulting in the final attention-enhanced output \( H^A \). This output incorporates both the original features and the selectively weighted information identified as most relevant by the attention mechanism. It is calculated as
\begin{align}
H^A = W_A (H \odot H^a) \in \mathbb{R}^{T \times d'},
\label{eq:attention_output}
\end{align}
where \( W_A \in \mathbb{R}^{d' \times d} \) and \(\odot\) denotes the Hadamard product (element-wise). Next, we fit \(H^A\) into the GAT to capture the spatial relationships, as explained below.

\section{Graph Attention Networks (GATs)}
To introduce GATs, we begin by discussing graphs, a data structure for modeling network topology. Graphs offer a natural representation for non-Euclidean structured data, such as communication networks.  A graph is defined as \( G = (V, E) \), where \( V \) represents nodes (e.g., routers) and \( E \) represents edges (e.g., links). The connectivity of the graph is encoded in the adjacency matrix \( A \), where \( A_{ij} = 1 \) if nodes \( i \) and \( j \) are directly connected, otherwise \( A_{ij} = 0 \). Although traditional graph representations are \textit{static}, meaning that the connectivity between nodes, encoded in the adjacency matrix \( A \), is fixed, therefore once an edge \( (i, j) \in E \) is defined (that is, \( A_{ij} = 1 \)), it is treated as equally important at all times, regardless of their dynamic factors such as traffic congestion or link delays. To address this, Graph Attention Networks (GATs) introduce a mechanism that allows each node to dynamically weigh their neighbors through the use of attention weights, which is discussed below. 

First, we construct feature representations for all nodes in the network. For each node \( i \), we define a feature vector \( x_i \in \mathbb{R}^K \), which may include information such as a snapshot or even a time series of delay at the node, or the delay on its adjacent links. Collectively, we denote the feature matrix as \( \mathbf{x} = \{ x_1, x_2, \ldots, x_N \} \in \mathbb{R}^{N \times K} \), where \( N \) is the total number of nodes and \( K \) is the dimensionality of the feature space. To capture the relationship or relative importance between a node \( i \) and its neighbor \( j \), the GATs calculate an attention coefficient \( e_{ij} \) for each pair of nodes as follows:
\begin{equation}
e_{ij} = \text{LeakyReLU} \left( \mathbf{a}^T \left[ \mathbf{W} x_i \, \| \, \mathbf{W} x_j \right] \right),
\label{eq:gat_attention}
\end{equation}
where the input features \( h_i \) and \( h_j \) are linearly transformed using a shared learnable weight matrix \( \mathbf{W} \in \mathbb{R}^{K' \times K} \), and \( \| \) denotes vector concatenation, and \( \mathbf{a} \in \mathbb{R}^{2K'} \) is a learnable weight vector that scores the relevance of each neighbor pair. The LeakyReLU activation is used to encourage sparsity in the learned attention weights, where only neighbors with strongly influence will be assigned higher attention scores, while less relevant neighbors are still down-weighted, but not ignored entirely. 

Since \( e_{ij} \) are raw, unnormalized attention scores, we need to transform them into normalized coefficients that sum to one over all neighbors of node \( i \). We achieve this using the softmax function:
\begin{equation}
\alpha_{ij} = \frac{\exp(e_{ij})}{\sum_{k \in \mathcal{N}_i} \exp(e_{ik})},
\label{eq:gat_normalize}
\end{equation}
where \( \mathcal{N}_i \) denotes the set of neighbors of node \( i \). As a result, \( \alpha_{ij} \) represents the proportion of attention that node \( i \) allocates to its neighbor \( j \).

After obtaining \( \alpha_{ij} \), we next aggregate the neighbors’ information to update the representation for node \( i \). This is done by computing a weighted sum of the neighbor features weighted by their respective \( \alpha_{ij} \) as
\begin{equation}
\hat{x}_i = \sigma \left( \sum_{j \in \mathcal{N}_i} \alpha_{ij} \, \mathbf{W} x_j \right),
\label{eq:gat_aggregation}
\end{equation}
where \( \sigma(\cdot) \) is a non-linear activation function, such as ReLU or ELU. This allows node \( i \) to selectively integrate information from its neighbors based on their importance. We collect the updated features for each node in the following feature matrix:
\begin{equation}
\mathbf{\hat{x}} = \{{\hat{x}}_1, {\hat{x}}_2, \ldots, {\hat{x}}_N\} \in \mathbb{R}^{N \times K}.
\label{eq:gat_output}
\end{equation}
Next, we describe how the agent interacts with the MDP-modeled environment and makes adaptive routing decisions by taking advantage of the GRU, temporal attention, and GAT modules discussed above.

\section{STRL Algorithm}
\subsection{MDP Framework}
\label{sec:MDP}
First, we model the routing optimization problem as a Markov Decision Process (MDP). The state represents all information available to the agent at each decision step. Specifically, at each time step \(t\), we define the state \(\mathcal{S}(t)\) as the past \(T\) time steps of node delays and outgoing link delays for all nodes. Specifically, for each node \(i\), let \(D_i^t = [d_i(t-T),\, d_i(t-T+1),\, \dots,\, d_i(t)]\) denote its delay sequence, where \(d_i(t)\) is the delay of node \(i\) at time \(t\). For each outgoing link \(k\) of node \(i\), let \(L_{i,k}^t = [l_{i,k}(t-T),\, \dots,\, l_{i,k}(t)]\) denote the delay sequence, where \(l_{i,k}(t)\) is the delay of the \(k\)th outgoing link at time \(t\). The state is thus:
\begin{align}
\mathcal{S}(t) = \{\, D_i^t,\, L_{i,1}^t,\, \dots,\, L_{i,K}^t\ |\ i=1,\dots,N\,\} \in \mathbb{R}^{N \times (K+1) \times (T+1)},
\label{eq:state}
\end{align}
where \( K \) is the maximum number of outgoing links among all nodes. For nodes with fewer than \( K \) outgoing links, the corresponding link delay sequences are padded with zeros. 

Next, we define the action space \(\mathcal{A}\) as the set of decisions the agent makes based on the observed state. At each time step \(t\), the agent outputs an action vector \(A^t = [a_1, \dots, a_N]\), where each \(a_i\) is a score for node \(i\) that measures how efficient that node is. Using these scores, we run Dijkstra’s algorithm to find the top \(k\) paths from the source to the destination with the highest cumulative scores. We then use the selected path to route next round simulation of \(M\) traffic.

After the agent selects an action, the environment responds with a reward. We define the reward at each time step as the ratio of system throughput to average end-to-end delay. The average end-to-end delay is the mean delay of all packets successfully transmitted from source to destination during the last simulation round. System throughput is defined as the proportion of traffic injected into the network by its source that reaches its destination within the simulation. 

\subsection{ST Architecture}
Now we discuss how the agent makes an action. First, the agent inputs the state \(\mathcal{S}(t)\) into GRU, going through the gating mechanism to discard old and retain new information via Equations~(\ref{eq:gating_reset_gate}), (\ref{eq:hidden_layer_output}), (\ref{eq:unit_output}), then output the hidden state \(H \in \mathbb{R}^{T \times d}\) in (\ref{eq:hidden_state}) that maintains only relevant memory. 

Next, to improve the model ability to capture long-range dependencies and mitigate the vanishing gradient problem, we apply a temporal attention mechanism to the hidden states $H$ generated by the GRU. Specifically, we first construct the query, key, and value representations from $H$ as described in Equation~(\ref{eq:self_attention_components}). For each time step $i$, we compute the attention weights that quantify the relevance of every past time step $j$ to $i$, following Equations~(\ref{eq:attention_weights})--(\ref{eq:attention_matrix}). This results in a new representation $H^A \in \mathbb{R}^{T \times d'}$, as defined in Equation~(\ref{eq:attention_output}), where each row aggregates information from all previous time steps according to their learned importance.

Next, we reshape $H^A$ into a per-node feature matrix of dimension $N \times K$, where $K = \frac{T \times d}{N}$, such that each node is associated with a feature vector of length $K$. Intuitively, this means we are taking all the memory the GRU has collected across time and distributing it evenly across all nodes, so each node receives its own feature vector that captures both recent history and the current state. 

Then, to model spatial relationships, we compute the attention coefficient $e_{ij}$ between each node $i$ and its neighbor $j \in \mathcal{N}_i$ as defined in Equation~(\ref{eq:gat_attention}). After normalizing these coefficients according to Equation~(\ref{eq:gat_normalize}), we update feature of each node by aggregating features of its neighbors weighted by their spatial attention, as in Equation~(\ref{eq:gat_aggregation}). The GAT layer thus produces updated node features $\hat{x} \in \mathbb{R}^{M \times N}$, which can be thought of as a rich summary of what that node has learned from both its recent history and from its neighbors in the network.

\begin{figure}[htbp]
    \centering
    \vspace{-1ex}
    \includegraphics[width=0.8\textwidth]{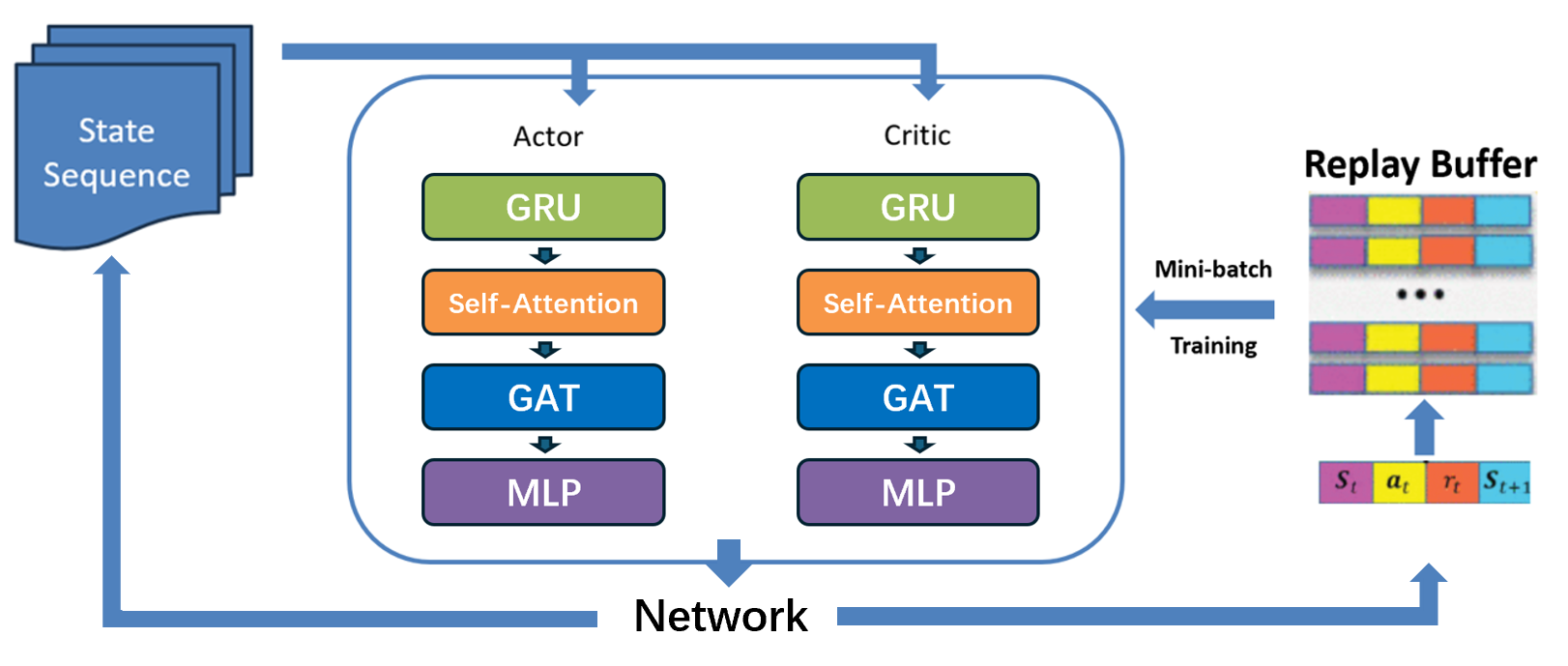}
    \caption{STRL Architecture for Solving NSFNet Topology}
    \label{fig:model_architecture}
\end{figure}

Finally, we generate the action vector $A^t$ from $\hat{x}$ by passing it through a Multi-Layer Perceptron (MLP) head. The MLP consists of a sequence of fully connected layers with Leaky ReLU activations and dropout for regularization, along with the layer normalization to improve gradient flow. Notably, the MLP is constructed to preserve the dimensionality of $\hat{x}$, so the output remains of size $N$ and is used as the action, where each value represents the efficiency score for each node. The purpose of the MLP is to allow the model to learn complex, non-linear mappings from each node's feature representation to its action, which will be used to influence the environment. The overall architecture is illustrated in Fig~\ref{fig:model_architecture}.

\subsection{RL Agent Training}
\label{sec:agent_training}
Using the action vector generated by the agent, where each value represents the efficiency score of a node, we enumerate all possible paths from the source to the target node. For each path, we calculate the total efficiency score by summing the scores of all nodes along the path. Dijkstra’s algorithm is then applied to identify the top \(k\) paths with the highest aggregate scores, which are deemed the most efficient for routing the traffic in the next time step, with each time step processing \(P\) incoming packets. After each time step, we store experiences in the form of (state, action, reward) tuples in a replay buffer, from which a batch of \(M\) experiences is sampled to compute the loss as defined in Eq.~\ref{eq:loss}. Model parameters are subsequently updated using the Bellman equations and policy gradient methods detailed in Section~\ref{sec:DDPG}.

\section{Application of STRL to Network Routing}
\subsection{Topology and Dataset}
\label{subsec:topo}
\begin{figure}[htbp]
    \centering
    \includegraphics[width=0.6\textwidth]{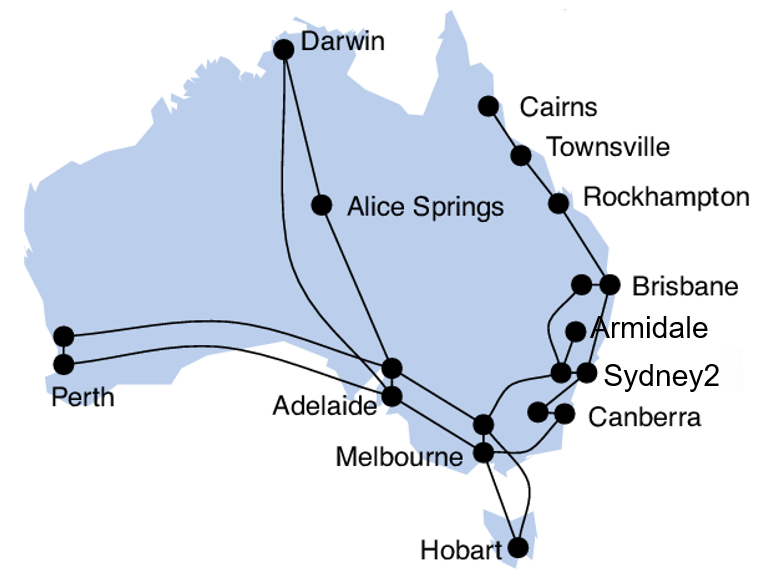}
    \caption{AARNet}
    \label{fig:aarnet}
\end{figure}

\begin{figure}[htbp]
    \centering
    \includegraphics[width=0.8\textwidth]{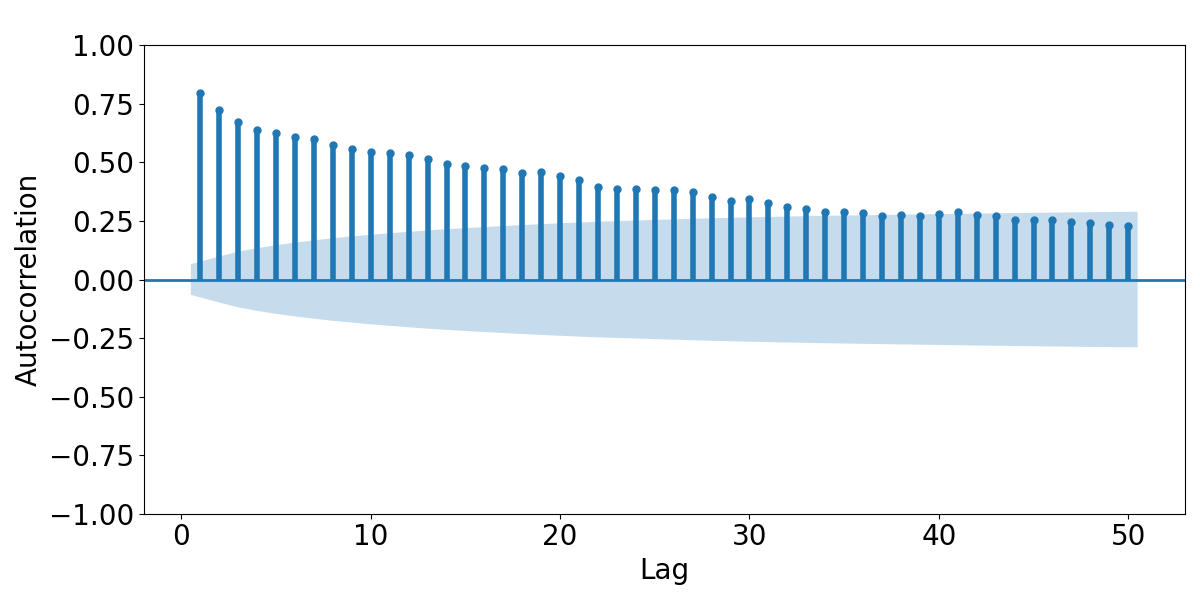}
    \caption{Autocorrelation Function (ACF) of Packet Arrival Rate over 15 minutes with Time Lag in Seconds}
    \label{fig:acf}
\end{figure}

Our experiment is based on the AARNet (Australian Academic and Research Network) topology from the Internet Topology Zoo, consisting of 17 nodes and 27 edges (see Fig.~\ref{fig:aarnet}). Each node represents a Point of Presence (PoP), and each edge denotes a network link between PoPs. For traffic data, we use real-world Internet traces from MAWI (Measurement and Analysis on the WIDE Internet), specifically the Samplepoint-F trace recorded on April 5, 2025, at 14:00 JST, spanning 15 minutes. Each entry in the trace corresponds to a packet and we aggregate the data to obtain packet arrival rates (in terms of the number of packets per second) during this time period. 
To examine temporal patterns in the traffic, we conduct an autocorrelation analysis on the packet arrival rates for the last 100 seconds (see Fig.~\ref{fig:acf}). Autocorrelation measures how current values in a time series relate to their past values at different time lags. Here, the x-axis represents the time lag in seconds, and the y-axis shows the autocorrelation coefficient with the shaded area being the 95\% confidence interval under the null hypothesis of zero correlation. We find that autocorrelation remains significantly positive for lags up to 40 seconds, meaning that arrival rates from the past 40 seconds are predictive of the current arrival rate. This confirms the non-Markovian pattern of packet arrivals, motivating us to include the past 40 seconds of traffic data as input to the RL agent. In the next section, we discuss how these temporal dependencies are represented in our simulation for the network environment.
\subsection{Network Enviornment}
\label{sec:system_model}
We first introduce the processing mechanism at the node and link levels. Let \( \lambda(t) \) be the number of packets per second, and we inject \( \lambda(t) \) into the AARNet. Each node \(i\) receives a continuous stream of traffic and processes it at a rate \( \mu_i \) (i.e., the number of packets being processed per second). The processing time at each node is then determined as \(\lambda(t)/\mu_i\). If the node is busy, the incoming traffic will be stored in the node's buffer. 
After processing, each node will forward the packets to an outgoing link. If the link is busy, the packets will be queued in the link buffer. Each link then obtains the transmission time \(T(t)\) based on the bandwidth demanded for the incoming traffic \(R(t)\), calculated as the product of the average packet size \( L \) and the number of incoming packets per second, and its own capacity \(B\):
\[
T(t) = \frac{R(t)}{B}, \quad R(t) = \lambda(t) \cdot L.
\]
Then, each link forwards packets to the connected node and repeats the same processing as described. Note that the time interval between successive traffic arrivals is set to one second to align with the non-stationary patterns in packet arrivals in seconds in Fig.~\ref{fig:acf}. 

Next, we describe how the environment is initialized for agent interaction. We begin with a warm-up phase in which traffic is injected between all possible node pairs, and forwarding paths are selected at random. This warm-up continues until the network reaches a steady state, where all nodes and links are sufficiently utilized. Following the warm-up, we run 1,000 training episodes, each consisting of 100 time steps. At each step, the agent interacts with the environment by observing the current state, taking actions, and receiving rewards, as described in Section~\ref{sec:agent_training}. The key parameter values for the agent, environment, and training process are summarized in Table~\ref{tab:strl_summary}. 

\label{sec:apply}

\section{Performance Evaluation}
\subsection{Baseline Models}

We compare our STRL agent with two baselines: Spatial RL (SRL) and Temporal RL (TRL). SRL shares the same architecture as STRL but excludes the GRU layers to disable it from capturing temporal patterns in the traffic. Similarly, TRL shares the same architecture as STRL but excludes the GAT layers, so the TRL architecture cannot see spatial relationships between nodes. This comparison aims to show how missing either temporal patterns or spatial relationships can impact routing performance. We also assess the robustness of these approaches to network topology changes.

\subsection{Results}
\label{sec:results}
\begin{figure}[h!]
    \centering
    \vspace{-1ex}
\includegraphics[width=0.8\textwidth]{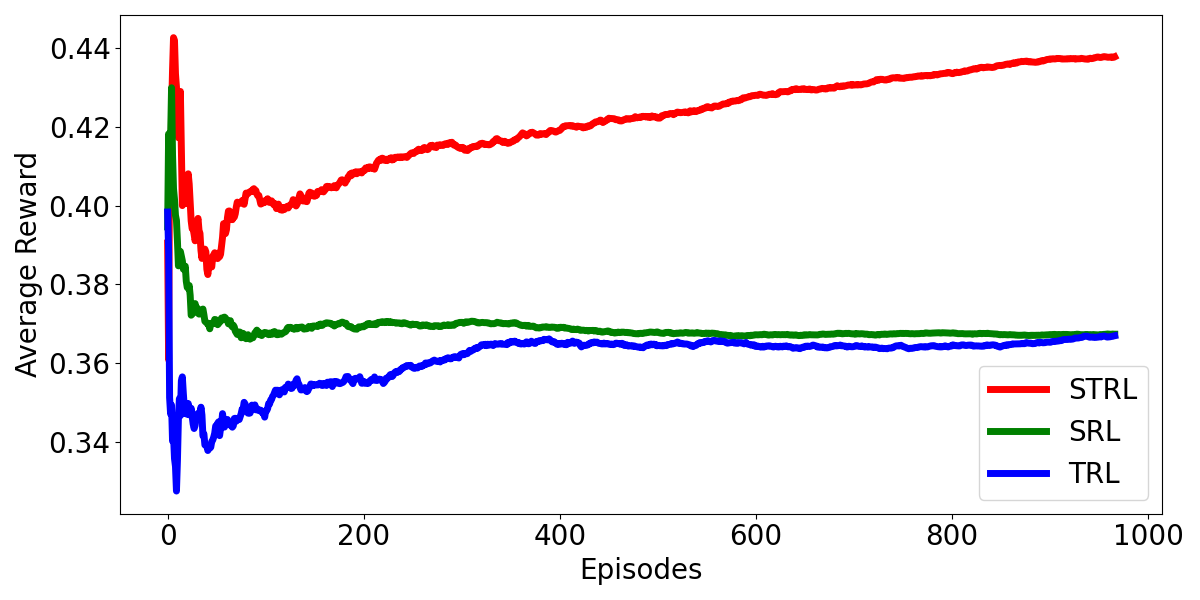}
    \caption{Comparison of average episode rewards across STRL, STL, and TRL}
    \label{fig:reward_total}
\end{figure}

As we can see from Fig~\ref{fig:reward_total}, STRL outperforms TRL and SRL by 19.20\% and 19.33\%, respectively, in reward, which is defined earlier as the ratio of the system throughput to the average end-to-end delay. This suggests the importance of considering \textit{both} spatial relationships in network topology and temporal dependencies in Internet traffic for making efficient routing decisions.

\begin{figure}[h!]
    \centering
    \vspace{-1ex}
\includegraphics[width=0.8\textwidth]{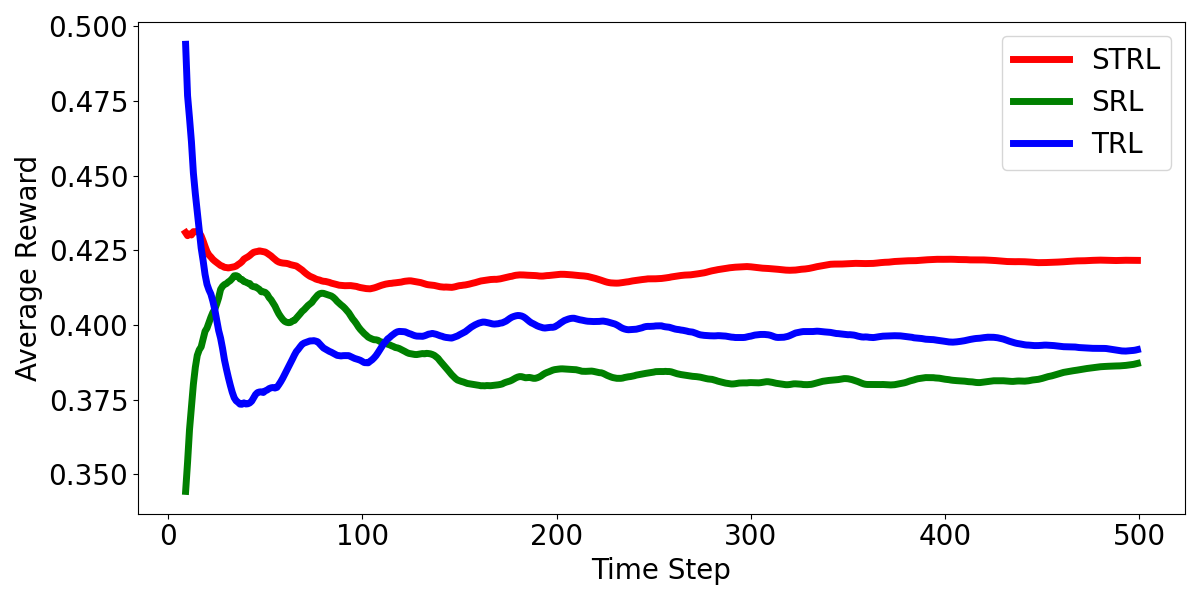}
    \caption{Comparison of average episode rewards under topology change}
    \label{fig:reward_topo}
\end{figure}

Next, we introduce a topology change to the network and see how it affects their performance. Initially, the node Armidale is a leaf node connected only to Sydney2. We modify the topology by adding two new links to connect Armidale with Rockhampton and Townsville. This will change the spatial relationships around Armidale and also cause the temporal patterns to change at these nodes. After 500 time steps of inference, the resulting performance is shown in Fig.~\ref{fig:reward_topo}, where STRL outperforms SRL and TRL by 8.65\% and 7.40\%, respectively. SRL can adapt quite quickly to this topology change due to its ability to access spatial relationships through the GAT layers, whereas TRL experiences a significant drop in performance. STRL, which combines both spatial and temporal information, can adapt with better performance in presence of the topological change when compared with the baseline approaches, SRL and TRL.

\begin{figure}[htbp]
    \centering
    \includegraphics[width=0.8\textwidth]{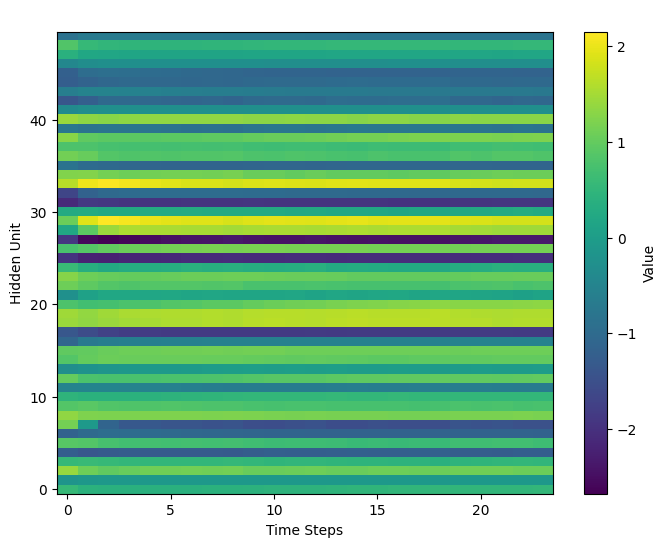}
    \caption{GRU Hidden State Heatmap}
    \label{fig:hidden_state}
\end{figure}

Next, we illustrate how STRL leverages non-Markovian traffic patterns in its decision-making by visualizing the GRU hidden states \( H \). Each hidden unit encodes different aspects of temporal information at each time step. Our architecture employs 40 time steps with 95 hidden units per step; however, for clarity, we only display the first 50 units over the first 25 time steps in Fig.~\ref{fig:hidden_state}.

The visualization shows how individual units in the y-axis capture various temporal dependencies for each past time step in the x-axis. For instance, the \(18^{\text{th}}\) and \(19^{\text{th}}\) units maintain consistently similar values across all time steps, suggesting that information from the entire 25-step window is uniformly important for their predictions at the next time step. Other hidden units may capture short-term dependencies. For example, the \(7^{\text{th}}\) hidden unit places greater importance on the most recent one or two time steps, as indicated by higher values at the beginning of the time window compared to the later time steps.


\section{Conclusion}
This paper has proposed a spatial-temporal reinforcement learning (STRL) approach for packet routing with non-Markovian traffic in communication networks. The proposed architecture outperforms in terms of training performance and robustness when compared with the baseline approaches capturing only spatial or temporal dynamics. Several improvements can be made to the proposed STRL approach. For example, although a self-attention mechanism is employed in the STRL, future work can investigate transformer-based models \cite{vaswani2017attention} to better capture long-range dependencies in sequential traffic data. Additionally, in large-scale networks, a distributed approach should be considered to address the scalability issues.

\section{Appendix}
\begin{table}[H]
\centering
\caption{Summary of Hyperparameters}
\label{tab:strl_summary}
\begin{tabular}{ll}
\hline
\textbf{Notation} & \textbf{Description / Value} \\
\hline
$T$ & Time window: 40\\
$F$ & Number of features in the GRU input: 5 \\
$d$ & GRU hidden dimension: 95 \\
$d_q$ & Latent dimension for query vector: 95 \\
$d_k$ & Latent dimension for key vector: 95 \\
$d_v$ & Latent dimension for value vector: 95 \\
$d'$ & Output dimension for temporal attention mechanism: 95 \\
$N$ & Number of nodes: 17 \\
$K$ & GAT input channels per node: 5 \\
$K'$ & GAT output channels per node: 5 \\
$\phi$ & Hidden dimension in MLP: 512 \\
$\eta_{\mu}$ & Learning rate for acton: 0.001 \\
$\eta_{Q}$ & Learning rate for critic: 0.001 \\
$\gamma$ & Discount factor: 0.6 \\
$\rho$ & Soft target update coefficient: 0.2 \\
$\epsilon$ & Initial exploration rate: 0.5 \\
$k$ & Top-$k$ paths per source-destination pair: 5 \\
$M$ & Batch size: 32 \\
$\delta$ & Dropout rate: 0.5 \\
$P$ & Number of incoming packets per second: 233233 \\
$B$ & Link capacity: 1000 Mbps \\
\hline
\end{tabular}
\end{table}



\end{document}